\documentclass{article}
\usepackage{amssymb}
\usepackage{xcolor}
\usepackage{graphicx}
\usepackage{courier}

\title{{\bf An Empirical Comparison of Three Inference Methods}}

\author{{\bf David Heckerman}\\
Medical Computer Science Group\\ 
Knowledge Systems Laboratory\\
Stanford University\\ 
Stanford, California 94305}

\date{}

\begin{document}

\maketitle

\begin{abstract}
\noindent In this paper, an empirical evaluation of three inference methods for
uncertain reasoning is presented in the context of Pathfinder, a large
expert system for the diagnosis of lymph node pathology.  The
inference procedures evaluated are (1) Bayes' theorem, assuming
evidence is conditionally independent given each hypothesis, (2)
odds--likelihood updating, assuming evidence is conditionally
independent given each hypothesis and given the negation of each
hypothesis, and (3) a inference method related to the Dempster--Shafer
theory of belief.  A decision-theoretic approach is introduced for
evaluating the performance of expert systems.  This approach, when
combined with a more traditional expert-rating method for evaluation,
provides insights about various components of the inference process.
\end{abstract}

\section{Introduction}

Several years ago, before learning much about methods for reasoning
with uncertainty, I and my colleagues began work on a large expert system,
called Pathfinder, that assists community pathologists with the diagnosis
of lymph node pathology.  Because the
Dempster--Shafer theory of belief was quite popular in our research group at the
time, we developed a inference method for our expert system inspired by
this theory.  The program performed fairly well in the opinion of the
expert pathologist who provided the knowledge for the system.

In the months following the initial development of Pathfinder,
several of us in the research group began exploring other methods for
reasoning under uncertainty.  We identified the Bayesian approach as a
candidate for a new inference procedure.  We realized that the
measures of uncertainty we assessed from the expert could be
interpreted as probabilities and we implemented a new inference method---
a special case of Bayes' theorem.

During this time, the expert was running cases through the program to test the
system's diagnostic performance.  One day, without telling him, we changed the
inference procedure to the Bayesian approach.  After running several cases with the
new approach, the expert exclaimed, ``What did you do to the program?  This is
fantastic!''

This experience was and still is in sharp conflict with the beliefs of many
researchers in the artificial-intelligence community.  At each of the first  three
AAAI uncertainty workshops, one or more researchers argued that the particular
inference method used does not significantly affect performance, at least in the
context of large real-world systems.  In this paper, a formal evaluation of the
performance of several inference methods is presented that confirms our early
experience with Pathfinder and refutes the claim made at the workshops.  Moreover,
it will be shown that the  Bayesian approach yields performance 
superior to that obtained with the other approaches in the domain of lymph-node
pathology.

In addition to describing the comparison, a new approach for evaluating the
performance of expert systems will be introduced.  This method, based in
decision-theory,  compliments a more traditional expert-rating approach to system
evaluation.  Both the new and traditional approaches will be used in the
experimental comparison of the inference procedures.

\section{The Domain}

AI researchers working on uncertain reasoning often complain that
the merits of one inference method versus those of another are evaluated
on the basis of only theoretical considerations.
Another complaint is that evaluations
of performance are limited to small or artificial domains.  This
study is designed to address both of these complaints.  The Pathfinder
program reasons about virtually {\em all} diseases that occur
in a human lymph node (24 benign diseases, 9 Hodgkin's
lymphomas, and 18 non-Hodgkin's lymphomas.)  In addition, the program
includes an exhaustive list of clues or features that can be used to
help determine a diagnosis.  Over 100 morphologic features or
patterns within a lymph node that can be easily recognized under a
microscope are represented.  The program also contains
over 30 features reflecting clinical, laboratory, immunological, and
molecular biological information that is useful in diagnosis.

Because this study focuses on only one domain, these results should
not be extrapolated to other domains.  All that will be demonstrated
is that the use of different inference methods {\em can} affect
performance in a real-world system.  Researchers interested in
learning more about the relative merits of different inference methods
are encouraged to begin similar investigations in other domains.

\section{The Inference Methods}

The three inference methods evaluated are (1) a special case of Bayes'
theorem, (2) an approach related to the parallel combination function
in the certainty-factor (CF) model \cite{Shortliffe75a}, and (3) a
method inspired by the Dempster--Shafer theory of belief
\cite{Shafer76}.  All three approaches take a set of observations and
produce a belief distribution over disease hypotheses based on the
same expert probability assessments.  However, the second two
approaches deviate significantly from probabilistic reasoning.

All three approachs share the assumption that the hypotheses
represented by the system are mutually exclusive and exhaustive.
Furthermore, all three approaches assume that the diagnostic features
are, in some sense, independent.  The exact nature of independence
varies from method to method and is discussed in detail in a later
section.  It should be noted that, during the development of
Pathfinder, obvious dependencies among features were eliminated by
clustering highly dependent features.  For example, a
pattern called {\em necrosis} is seen in many lymph node
diseases.  The size of necrosis (percent area of lymph node showing
this pattern) and the distribution of necrosis are two strongly
interrelated features, and both are important for diagnosis.  To remove the
dependency, a single feature ``necrosis size and distribution'' was
created which had the mutually exclusive and exhaustive values
``nonextensive and focal,'' ``nonextensive and multi focal,''
``extensive and focal,'' and ``extensive and multi focal.''  These values
were created by taking the cross-product of the values for individual
features pertaining to necrosis size and necrosis distribution.

Before describing the inference methods, some definitions and notation
are introduced. The mutually exclusive and exhaustive disease
hypotheses will be denoted by the symbol $d$ with a subscript---for
example, $d_{j}$.  Similarly, the symbol $f_{k}$ refers to the
$k$th feature in the knowledge base.  Each feature is associated
with a set of mutually exclusive and exhaustive {\em values}.  The
$i$th value of the $k$th feature is denoted by $v_{ki}$.  A
given feature and a value for that feature together constitute an
{\em observation}.  The term $f_{k}v_{ki}$ denotes an
observation of the $i$th value for the $k$th feature.  For the
sake of brevity, a set of observations 
$f_{1}v_{1i} \ldots f_{n}v_{ni}$ will be denoted by the symbol
$\xi$.
Finally, two conditional independence assumptions associated with the inference
procedures are introduced here for reference.  The first assumption is that
evidence is conditionally independent on disease hypotheses.
Formally, the assumption is that, for any combination of
observations \mbox{$f_{1}v_{1i} \ldots  f_{n}v_{ni}$},
\begin{equation}
p(f_{1}v_{1i} \ldots f_{n}v_{ni}|d_{j})  =  p(f_{1}v_{1i}|d_{j}) \ldots p(f_{
n}v_{ni}|d_{j}) \label{ciH}
\end{equation} 
The second assumption is that evidence is
conditionally independent on the {\em negation} of the hypothesis.
Specifically, for any combination of
observations \mbox{$f_{1}v_{1i} \ldots  f_{n}v_{ni}$},
\begin{equation}
p(f_{1}v_{1i} \ldots f_{n}v_{ni}|\overline{d_{j}})  =  p(f_{1}v_{1i}|\overline{d_{j}})
\ldots p(f_{n}v_{ni}|\overline{d_{j}}) \label{cinotH} 
\end{equation}  
Both Equations \ref{ciH} and \ref{cinotH} apply to each disease
hypothesis $d_{j}$.

\subsection{Simple Bayes Method}

The first inference method is Bayes' theorem under the assumption that
features are conditionally independent on the disease hypotheses
(Equation \ref{ciH}).  In particular, if observations 
\mbox{$\xi = f_{1}v_{1i} \ldots f_{n}v_{ni}$} are made, the
probability of the $j$th disease is given by
\begin{equation}
p(d_{j}|\xi)  =  \frac{p(d_{j}) p(f_{1}v_{1i}|d_{j}) \ldots p(f_{n}v_{
ni}|d_{j})}{\sum_{j} p(d_{j}) p(f_{1}v_{1i}|d_{j}) \ldots p(f_{n}v_{ni}|d_{
j})} \label{Bayes}
\end{equation} 
This inference procedure will be called the {\em simple Bayes method} to
emphasize the conditional independence assumptions it embodies.
Note that the only assessments required by this approach are
the probabilities $p(f_{k}v_{ki}|d_{j})$ for each
combination of $f_{k}$, $v_{ki}$, and $d_{j}$, and the
prior probabilities $p(d_{j})$ for each disease.  The other
two inference methods require the same assessments.

\subsection{Odds--Likelihood Method}

The second inference method begins with a form of Bayes' theorem under
the assumption that evidence is conditionally independent on both the
hypotheses and on the negation of the hypotheses (Equations \ref{ciH}
and \ref{cinotH}).  Under these assumptions, Bayes' theorem for the
$j$th disease given observations $\xi$ can be written
\begin{equation}
\frac{p(d_{j}|\xi)}{p(\overline{d_{j}}|\xi)}  =  \frac{p(d_{j})}{p(\overline{d_{j}})}
\frac{p(f_{1}v_{1i}|d_{j})}{p(f_{1}v_{1i}|\overline{d_{j}})} \cdots \frac{p(f_{n}v_{
ni}|d_{n})}{p(f_{n}v_{ni}|\overline{d_{n}})} \label{BU1}
\end{equation} 
The ratio on the left-hand side and the first ratio on the
right-hand side of Equation \ref{BU1} are the {\em posterior} and
{\em prior} odds of $d_{j}$, respectively. In general, the odds of
an event is just a simple monotonic transformation of the probability
of the event, given by
\begin{displaymath}
O = \frac{p}{1-p}
\end{displaymath} 
The remaining terms of Equation \ref{BU1} are called likelihood ratios.
As can be seen from Equation \ref{BU1}, the likelihood ratio
$p(f_{k}v_{ki}|d_{j})/p(f_{k}v_{ki}|\overline{d_{j}})$
is a measure of the degree to which observing feature value
$f_{k}v_{ki}$ updates or changes the degree of belief in disease
hypothesis $d_{j}$.

In the version of this inference method evaluated in this paper, the
likelihood ratios are not assessed directly.  Instead, the numerator,
$p(f_{k}v_{ki}|d_{j})$ is assessed directly, and the
denominator, $p(f_{k}v_{ki}|\overline{d_{j}})$, is computed using
\begin{displaymath}
p(f_{k}v_{ki}|\overline{d_{j}})  =  \frac{p(f_{k}v_{ki}) - p(f_{k}v_{ki}|d_{j}) p(d_{
j})}{p(f_{k}v_{ki})}
\end{displaymath}
where
\begin{displaymath}
p(f_{k}v_{ki})  =  \sum_{j} p(f_{k}v_{ki}|d_{j}) p(d_{j})
\end{displaymath}
Thus, this inference method makes use of exactly the same
assessments as does the simple Bayes approach.  The likelihood ratios were not
assessed directly because the expert found that likelihood ratios were much
more difficult to assess than were conditional probabilities
$p(f_{k}v_{ki}|d_{j})$.  It would be interesting to conduct a
comparison similar to the one described in this paper using an expert
who is willing to assess likelihood ratios directly.

Johnson \cite{Johnson86a} has demonstrated that the conditional-independence
assumptions embodied in Equation \ref{BU1} typically are not
compatible with the updating of $n$ mutually exclusive and
exhaustive hypotheses, when $n$ is greater than two.  In particular,
he has shown that consistently updating more than two mutually
exclusive and exhaustive hypotheses under the conditional-independence
assumptions used to derive Equation \ref{BU1} is
possible only when each hypothesis is updated by at most one
observation.

In Pathfinder, this highly restrictive condition required for
consistent updating is not met.  Each disease hypothesis is updated
by many observations in the knowledge base.  As a result, 
Equation \ref{BU1} produces an inconsistent probability
distribution over diseases in which the posterior probabilities of
disease do not sum to one.  To circumvent this problem, the disease
probabilities are renormalized after Equation \ref{BU1} is applied to
the evidence.  This completes the description of the second
approach, which will be called the {\em odds--likelihood method}.  

It should be mentioned that the odds--likelihood approach is closely
related to the parallel combination function used in the CF model.
In fact, it was shown that the multiplicative combination of
likelihood ratios seen in Equation \ref{BU1} maps exactly to the
parallel combination function when a certainty factor is identified
with a simple monotonic transformation of the likelihood ratio
\cite{Heckerman86b}.  Moreover, in MYCIN---the expert system for which
the CF model was designed---certainty factors of
mutually exclusive and exhaustive sets of hypotheses are renormalized
to sum to unity \cite{Shortliffe74}.  This form of renormalization
does not correspond directly to the renormalization of probabilities
in the second inference method, but it is similar in spirit.

\subsection{Naive Dempster--Shafer Method}

The third inference method has an interesting history. It was
developed by researchers, including myself, who at the time knew
little about methods for uncertain reasoning.  As the method was primarily
motivated by the Dempster--Shafer theory of belief, it will be called
the {\em naive Dempster--Shafer method}.  It should be emphasized that
the approach is fraught with difficulties, some of which will be
addressed in the Discussion section.  Perhaps the
exposition will serve as a warning to the uncertainty in
AI community as to what can happen when a group of novice researchers
attempts to cope with the conflicting uncertainty literature.

As members of a medical information-science group, we were familiar
with the inference method used by INTERNIST-1, an expert system for the
diagnosis of disease across all diseases in internal medicine
\cite{Miller82}.  The inference procedure used by INTERNIST-1
incorporates two measures of uncertainty, an {\em evoking
strength} and a {\em frequency}.  An evoking strength for disease
$d_{j}$ and observation $f_{k}v_{ki}$, denoted
$ES(d_{j},f_{k}v_{ki})$, represents the degree to which the
observation ``evokes'' or ``confirms'' the disease \cite{Miller82}.  In
contrast, a frequency for disease $d_{j}$ and observation
$f_{k}v_{ki}$, denoted $FQ(d_{j},f_{k}v_{ki})$,
represents the ``likelihood'' of an observation given the disease
\cite{Miller82}.

Because we initially had planned to use the INTERNIST-1 inference
procedure, our expert assessed both an evoking strength and a frequency
(on a continuous scale from 0 to 1) for each disease--observation
pair.  Before we began programming the approach, however, several members
of our group argued that a more principled approach
should be used to combine the measures of confirmation we had assessed.  In
particular, they argued that the Dempster--Shafer theory of belief
should be used to combine evoking strengths.

After exploring of the Dempster--Shafer theory, we decided to
construct a separate frame of discernment,
\mbox{$\theta_{j} = \{d_{j},\overline{d_{j}}\}$},
for each disease hypothesis $d_{j}$.  In
this framework, the evoking strength for a disease--observation pair is
interpreted as a mass assignment to the singleton disease
hypothesis:
\begin{equation}
m_{f_{k}v_{ki}}(\{d_{j}\}) = ES(d_{j}, f_{k}v_{ki}) \label{ES1}
\end{equation} 
The remainder of the mass,
$1 - ES(d_{j},f_{k}v_{ki})$, is assigned to $\theta$.
Mass assignments of this form follow the approach taken by Barnett
\cite{Barnett81}.  With this interpretation, Dempster's rule of
combination can be used to determine the mass assigned to the
singleton hypothesis $\{d_{j}\}$, given observations $\xi$.
In particular, Barnett showed that
\begin{equation}
m_{\xi}(\{d_{j}\})  =  1 - \prod_{k} \left(1 - m_{f_{k}v_{ki}}(\{d_{j}\})\right)
\label{ES2} 
\end{equation}  
In this framework of simple belief functions,
the mass assigned to the singleton hypothesis $\{d_{j}\}$ is equal to the belief
in $d_{j}$, denoted $Bel(\{d_{j}\})$.  Thus, combining Equations
\ref{ES1} and \ref{ES2}, we can compute the belief in disease $d_{j}$
given observations $\xi$, using
\begin{equation}
Bel_{\xi}(\{d_{j}\})  =  1 - \prod_{k} (1 - ES(d_{j},f_{k}v_{ki})) \label{ES3}
\end{equation} 
This inference method produces a number between zero and one for each
disease hypothesis.

At first, we ignored the frequencies provided by our expert.
However, our expert kept insisting that his assessments of frequency
were much more reliable than were his assessments of evoking strength.
This led us to study the INTERNIST-1 inference method more
carefully.  It became clear to us and to our expert that
the assessed evoking strengths were closely related to the posterior
probability of disease given an observation.  Also, it became
apparent that the assessed frequencies corresponded to the probability of an
observation given a disease.  Thus, we discarded the directly
assessed evoking strengths and replaced them with the calculated
values
\begin{eqnarray}
ES(d_{j},f_{k}v_{ki}) & = & p(d_{j}|f_{k}v_{ki}) \nonumber \\
 & = & \frac{p(d_{j}) p(f_{k}v_{ki}|d_{j})}{\sum_{j} p(d_{j}) p(f_{k}v_{ki}|d_{
j})}  \label{ES4}
\end{eqnarray}  
where each probability assessment $p(f_{k}v_{ki}|d_{j})$ is given by the frequency
$FQ(d_{j},f_{k}v_{ki})$. Equation \ref{ES4} follows from Bayes' theorem and the
assumption that diseases are mutually exclusive.

Equations \ref{ES3} and \ref{ES4} together provide a method for computing the
posterior degree of belief in each disease hypothesis from the prior probabilities
of disease, $d_{j}$, and the probabilities of an observation given disease,
$p(f_{k}v_{ki}|d_{j})$.  These are the same assessments that are required by the
two approaches described previously.  It should be noted that the resulting belief
distribution rarely sums to unity.  This fact is not a problem conceptually,
because the evaluation metrics used in the experimental comparison do not
require probabilistic interpretations for the distributions.  Nonetheless, the
distributions produced by this inference method are renormalized to one, so that 
during the evaluation process, the expert is not able to recognize them immediately
as being nonprobabilistic.
					
Like the odds--likelihood approach, the naive Dempster-Shafer method is
related to parallel combination in the CF model.  In fact, Equation \ref{ES2}
is exactly the parallel combination function for positive or
confirming evidence if certainty factors are identified with
singleton mass assignments.

\section{The Evaluation Procedure}

\begin{figure}
\begin{center}
\leavevmode
\includegraphics[width=3.0in]{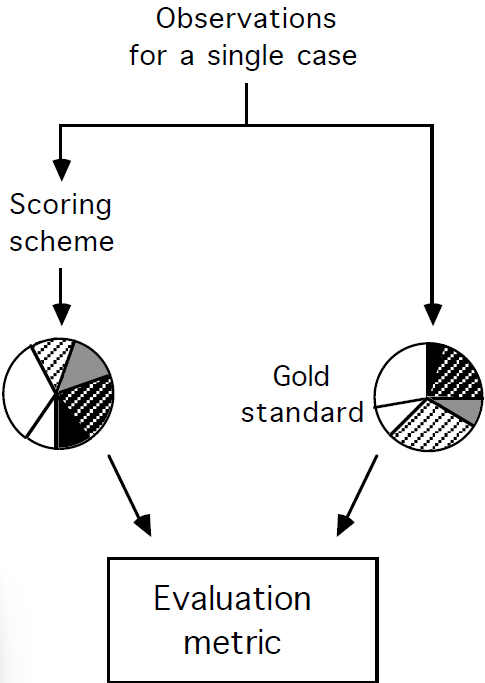}
\end{center}
\caption{An schematic of the evaluation procedure.}
\label{fevalf1}
\end{figure}

The procedure for evaluating the inference methods is outlined in
Figure \ref{fevalf1}.  Observations describing a lymph-node biopsy
of a patient are presented to an inference method.  The inference
method, in turn, produces a belief distribution over the disease
hypotheses.  Finally, the belief distribution is compared with the
{\em gold standard} probability distribution using an {\em evaluation
metric}.  The process is repeated for each of the three inference
methods.  In this section, the gold standard and evaluation metrics
are defined.

An important consideration underlying the definition of the gold
standard is the distinction between good {\em decisions} and good {\em
  outcomes}.  A good decision is one that is consistent with the
preferences and knowledge of a decision maker.  A good outcome is one
that is desirable to the decision maker.  Sometimes, a good decision
will, through a course of bad luck, lead to a bad outcome.
Conversely, a bad decision may, with good luck, lead to a good
outcome.  However, the best way to achieve good outcomes in the long
run, short of being all-knowing, is to make good decisions
consistently.  Therefore, the gold standards developed in this study
are designed to identify good decisions, not necessarily good
outcomes. This distinction between good decisions and outcomes is
recognized in several previous evaluations including the validation
experiments of Yu \cite{yu79a} \cite{yu79b}, Cooper \cite{Cooper84},
and Wise \cite{Wise86b}.

Two gold standards were used to compare the inference methods.  The first was
derived from the probability distribution over diseases that our expert assessed
using the same list of observations that was presented to the inference methods. 
The second was similar, except that the expert also reviewed the belief
distributions generated by the three approaches.  

The two gold standards each have advantages and disadvantages. Both isolate the
evaluation of the inference methods from actual outcomes.  The first gold standard
is useful because it serves to measure how well each inference method matches the
unaided reasoning of the expert. The second gold standard is useful because
individuals, including experts, often make mistakes when reasoning under
uncertainty in the sense that they violate highly desired principles of reasoning
\cite{Tversky74}.  Indeed, the terms {\em descriptive} and {\em normative} often
are used to distinguish how individuals actually reason and how they should
reason.  Of course, our expert is unlikely to appreciate his errors in reasoning,
and to adjust his assessments accordingly, simply by observing the output of the
three inference methods.  A decision analyst would argue, for example, that a
decision maker must make many iterations of the cycle comprising formulating
assumptions, assessing probabilities, and inspecting the consequences of the
assumptions and assessments before she can have any assurance that she is making a
good decision.  Such detailed iterations, however, are not possible in this
experimental comparison because the principles of reasoning underlying each approach
are not identical.\footnote{In fact, the principles underlying the odds--likelihood
and naive Dempster--Shafer approaches are unclear to the author.}  Developing a gold
standard corresponding to a good decision under the principles associated with one
of the inference methods would bias the results in favor of that inference method.
By allowing the expert to see the distributions generated by each approach, it is
only hoped that gross errors in reasoning, such as lack of attention to rare
hypotheses, will be reduced.  To emphasize the roles played by the first and second
gold standard, they will be called the {\em descriptive} and {\em
informed} gold standards, respectively.

\subsection{A Decision-Theoretic Evaluation Metric}

Two evaluation metrics are used to compare the inference methods.  One approach is
based on direct ratings given by the expert.  The order approach, described in this
section, is grounded in decision theory. Although other authors have suggested
similar approaches (for example, see Wise \cite{Wise86b}), the comparison described
in this paper is, to the knowledge of the author, the first to apply decision
theory to the evaluation of a large real-world expert system.

The fundamental notion underlying the decision-theoretic metric is
that some errors in diagnosis are more serious than others are.  For
example, if a patient has a viral infection and is incorrectly
diagnosed as having cat-scratch disease---a disease caused by an
organism that is killed with antibiotics---the consequences are not
severe.  In fact, the only nonnegligible consequence is that the
patient will take antibiotics unnecessarily for several weeks.
If, however, a patient has Hodgkin's disease and is incorrectly
diagnosed as having an insignificant benign disease such as a viral
infection, the consequences are often lethal.  If the diagnosis had
been made correctly, the patient would have immediately undergone
radio- and chemotherapy, with a 90-percent chance of a cure.
If the disease is diagnosed incorrectly, however, and thus is not treated,
it will progress.  By the time major symptoms of the disease appear and the patient
once again seeks help, the cure rate with appropriate treatment will have
dropped to leass than 20 percent.

A decision theoretic approach to evaluation recognizes such variation
in the consequences of misdiagnosis.  The significance of each possible
misdiagnosis is assessed separately.  More specifically, for each
combination of $d_{i}$ and $d_{j}$, a decision maker is asked,
``How undesirable is the situation in which you have disease
$d_{i}$ and are diagnosed as having disease $d_{j}$?''  The
disease $d_{j}$ is called the {\em diagnosis} and the preference
assessed is called the {\em diagnostic utility}, denoted
$U_{ij}$.  Details of the utility assessment procedure are
discussed in the following section.

\begin{figure}
\begin{center}
\leavevmode
\includegraphics[width=3.5in]{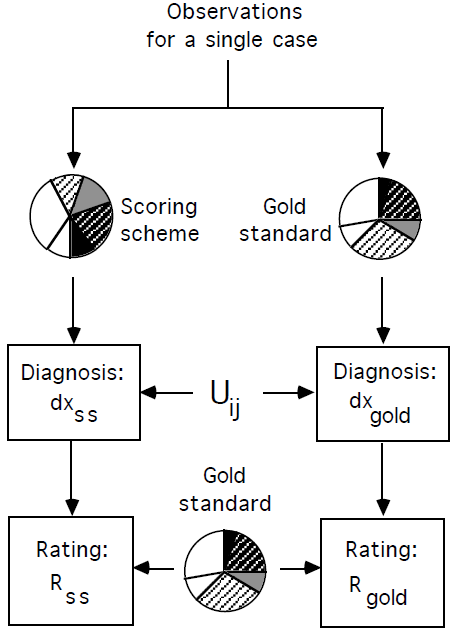}
\end{center}
\caption{The decision-theoretic evaluation procedure.}
\label{fevalf2}
\end{figure}

Once the diagnostic utilities are assessed, it is a straightforward
to evaluate each of the inference methods relative to the gold
standard.  The procedure for evaluation is shown in Figure
\ref{fevalf2}.  First, observations for a case are presented to a
inference method to produce a belief distribution over the disease
hypotheses, denoted $p_{ss}$.  In addition, the observations are
shown to the expert, who then assesses the gold-standard distributions,
denoted $p_{gold}$.  

Next, a {\em decision rule} is used to determine the optimal diagnosis
given each of the belief distributions.  In many systems that employ
methods for uncertain reasoning, a commonly used decision rule is to
choose the hypothesis with the highest degree of belief
\cite{deDombal72} \cite{Cleckner85}.  Formally, the optimal diagnosis 
$d_{ss}$ for a belief distribution $p_{ss}$ is given by
\begin{equation}
dx_{ss}  =  \arg\max_{i} \left[ p_{ss}(d_{i}) \right]  \label{eval3}
\end{equation} 
where $\arg\max_{i}$ returns the $d_{i}$ that maximizes the quantity
$p_{ss}(d_{i})$.  This rule for choosing the optimal diagnosis is  applied to  the
belief distributions produced by each of the inference methods.  Note that the rule
does not require that the degrees of belief computed by an inference procedure have
a probabilistic interpretation.

The gold-standard diagnoses are then determined.  The gold standards
are prescribed using a decision rule different from Equation \ref{eval3}.  In
particular, a gold-standard diagnosis is determined by finding the
diagnosis that maximizes the {\em expected utility} of the patient.  More
formally, 
\begin{equation}
dx_{gold}  =  \arg\max_{j} \left[ \sum_{i} p_{gold}(d_{i}) U_{ij} \right]
\label{eval1} 
\end{equation}
where $dx_{gold}$ denotes a gold-standard diagnosis.  For comparison, both
Equations \ref{eval3} and \ref{eval1}, with $p_{gold}$ replaced by $p_{ss}$, are
applied to the simple Bayes approach.  The application of Equation \ref{eval1} to
the simple Bayes approach is justified because the inference method produces a
legitimate probability distribution over diseases.

After the gold-standard diagnoses are determined, ratings for the two
distributions can be computed.  In this decision-theoretic
framework, the natural choice for a rating is the {\em expected utility} of
each diagnosis, where expectation is dictated by the distributions
used to derive the gold standards.  That is,
\begin{displaymath}
R_{ss}  =  \sum_{i} p_{gold}(d_{i}) U_{i,dx_{ss}}
\end{displaymath}
and
\begin{displaymath}
R_{gold}  =  \sum_{i} p_{gold}(d_{i}) U_{i,dx_{gold}}
\end{displaymath}
where $R_{ss}$ and $R_{gold}$ denote the ratings for the
inference method and gold-standard diagnoses, respectively.  Note that
the two ratings can be different only when the diagnoses prescribed
by the two distributions $p_{ss}$ and $p_{gold}$ are not the
same.

\subsection{An Expert-Rating Evaluation Metric}

In addition to the decision-theoretic approach, an expert-rating method is used to
compare the inference methods.  For each probability distribution, the expert is
asked, ``On a scale from zero to ten--- zero being unacceptable and ten being
perfect---how accurately does the distribution reflect your beliefs?''  The ratings
given by the expert are compared using standard statistical techniques.  Note that
gold standards are not explicitly elicited in this approach.

The expert-rating metric is used for two reasons.  First, expert-rating approaches
have been used frequently in expert system evaluations.  (See, for example,
Cooper \cite{Cooper84}.)  Therefore, it is useful to compare the
approach with the decision-theoretic method introduced in this paper.  Second, the
expert-rating and decision-theoretic approaches evaluated different aspects of
performance and are complementary.

\section{Utility Assessment}

In this section, several important issues surrounding the assessment
of diagnostic utilities are addressed and details of the procedure
for assessment are described.

An important consideration in the assessment of diagnostic utilities
is that preferences will vary from one decision maker to
another.  For example, the diagnostic utilities of decision makers
faced with the results of a lymph-node biopsy are likely to be influenced
by theier age, sex, and state of health.
Consequently, the ratings produced
by the decision-theoretic metric are meaningful to an individual only
to the degree that their diagnostic utilities match those used in the
evaluation.  

For this experimental comparison, the utilities of the expert on the Pathfinder
project were used.  The expert was chosen for two practical reasons.
First, being an expert, he was reasonably familiar with many of the
ramifications of correct and incorrect diagnosis.  Second, a good
working relationship with him had been established during the
construction of Pathfinder.  In future experiments, it would be
useful to generate a utility model using an expert clinician who
might have better insight into the preferences of a ``typical'' patient
making a decision based on the results of a lymph-node biopsy.

It is interesting to note that our expert, because he is an expert, had biases that
made his initial preferences deviate significantly from those of a typical patient. 
For example,  many sets of diseases of the lymph node currently have identical
treatments and prognoses.  Nonetheless, experts like to distinguish diseases within
each of these sets, because doing so allows research in new treatments to progress. 
That is, experts often consider the value of their efforts to future patients. In
addition, experts generally suffer professional embarrassment when their diagnoses
are incorrect.  Also, experts are concerned about the legal liability associated
with misdiagnosis.  In an effort  to remove these biases, our expert was
specifically asked to ignore these attributes of utility.  He was asked to imagine
that he himself had a particular disease, and to assess the diagnostic
utilities accordingly.

Another important consideration in almost any medical decision problem is the wide
range of severities associated with outcomes.  As mentioned previously, one
misdiagnosis might lead to inappropriate antibiotic therapy, whereas another might
lead to almost certain death. How can preferences across such a wide range be
measured in common terms?  Early attempts to resolve this question were fraught with
paradoxes.  For example, in a linear willingness-to-pay approach, a decision maker
might be asked, ``How much would you have to be paid in order to accept a one in
ten-thousand chance of death?''  If the decision maker answered, say, one thousand
dollars, the approach would dictate that he would be willing to be killed for
ten million dollars.  Clearly, this is absurd.

Recently, Howard has constructed an approach that avoids many of the paradoxes of
earlier models \cite{Howard80}.  Like many of its predecessors, the model deals with
determining what an individual would have to be paid to assume some chance of death,
and what he would be willing to pay to avoid a given risk.  Also like many of its
predecessors, Howard's model shows that, for small risks of death (typically, $p <
0.001$), the amount someone would be willing to pay or would have to be paid to avoid
or to assume such a risk is linear in $p$.  That is, for small risks of death, an
individual acts like an expected-value decision maker with a finite value attached
to his life.  For significant risks of death, however, the model deviates strongly
from linearity.  For example, the model shows that there is a maximum probability of
death, beyond which an individual will accept no amount of money to risk that chance
of death.  Most people find this to be an intuitive result. \footnote{The result
ignores considerations of legacy.}

In this paper, the details of the model will not be presented.
For a discussion of the approach see
\cite{Howard81}.  Here, we need only to assume that willingness to
buy or sell {\em small} risks of death is linear in the probability of
death.  Given this assumption, preferences for minor to major outcomes can
be measured in a common unit, {\em the probability of immediate,
painless death that a person is willing to accept to avoid a given outcome and
to be once again healthy}.  The undesirability of major outcomes can be
assessed directly in these terms.  For example, a decision maker
might be asked, ``If you have Hodgkin's disease and are incorrectly
diagnosed as having a viral infection, what probability of
immediate, painless death would you be willing to accept to
avoid the situation and to be once again healthy?''  At
the other end of the spectrum, the undesirability of minor outcomes
can be assessed by willingness-to-pay questions, and can be translated, via
the linearity result, to the common unit of measurement.  For
example, a decision maker might be asked, ``How much would you be willing
to pay to avoid taking antibiotics for two weeks?''  If he answered
\$100, and if his small-risk value of life were \$100,000,000, then the
answer could be translated to a utility of a 1 in 1,000,000 chance of
death.

Thus, the only major task in assessing the $U_{ij}$, aside from making the
direct assessments themselves, is the determination of the
decision maker's small-risk value of life.  Howard proposes a model by
which this value can be computed from other assessments.  A
simple version of the model requires a decision maker to trade-off the
amount of resources he consumes during his lifetime and the length of
his lifetime, to characterize his ability to turn present cash into
future income, and to establish his attitude toward risk.  However,
our expert did not find it difficult to assess the small-risk value of
life directly.\footnote{Howard also has observed that the small-risk
value of life can be assessed directly \cite{HowardPC87}.}  When asked
what dollar amount he would be willing to pay to avoid chances of
death ranging from $1$ in $20$ to $1$ in $1000$, he was consistent with the linear
model to within a factor of 2, with a median small-risk value of life
equal to \$10,000,000.

Note that, with this utility model, the ratings $R_{ss}$ assigned
to the inference methods will have units ``probability of death.''  In
many cases, we shall see that the differences between ratings are
small in these units (on the order of 0.001).  Consequently, it is
useful to define a {\em micromort}, a one in one million
chance of death.  In these units, for example, a
decision maker with a small-risk value of life of \$10,000,000 should
be willing to buy and sell risks of death at the rate of \$10 per
micromort.  This unit of measurement is also useful because it helps to emphasize
that the linear relationship between risk of death and
willingness to pay holds for only small probabilities of death.

Another important consideration is the complexity of the utility assessment
procedure.  There are 51 diseases represented in Pathfinder.  The direct measurement
of the $U_{ij}$ therefore requires \mbox{$51^{2} = 2601$} assessments.  Clearly, the
measurment process would be tedious.  Thus, several steps were taken to reduce the
complexity of the task.  First, the expert was asked to establish sets of disease
hypotheses that have identical treatments and prognoses.  An example of such a set is
the collection of nine types of Hodgkin's diseases represented in Pathfinder.
Patients with any of the nine types receive the same treatment and have the same
prognosis.\footnote{Prognosis for these nine types of Hodgkin's disease is determined
by the clinical stage, not by the specific type of disease.}  The expert identified
26 such ``equivalence classes,'' reducing the number of direct utility assessments
required to \mbox{$26^{2} = 676$}.  

Next, the expert was asked to order the utilities $U_{ii}$---
he was asked to order the undesirability of having each disease
and being diagnosed correctly.  After he had completed this ranking,
he was asked to quantitate each $U_{ii}$ in the manner
described previously.  It should be noted that the ordering of the
$U_{ii}$ was modified significantly during this process.  About
halfway through the procedure, he exclaimed, ``The dollar is
forcing me to think very carefully!''  It would be interesting to determine
whether most people respond in this way.  The results of such a study
would be interesting, particularly to reseachers in qualitative reasoning.

Finally, the off-diagonal utilities were assessed.  For each disease, the expert was
asked to quantify the undesirability of having the disease and being diagnosed as
having a different disease.  First, he identified the most similar preexisting
assessment.  It was then a simple matter to identify the differences between the
current assessment and the preexisting assessment, and to modify the utility
appropriately.  For example, given a patient with the disease sinus hyperplasia, the
only difference between her being diagnosed correctly and her being diagnosed with
cat scratch disease is that, in the latter case, the patient would take unnecessary
antibiotics for several weeks.  The expert said that he would be willing to pay \$100
to avoid taking the antibiotics, so this value (converted to micromorts) was
subtracted from the utility of being correctly diagnosed with sinus hyperplasia.

\section{Details of the Experiment}

Whenever possible, the conditions of the experimental comparison were arranged to
mimic the conditions under which Pathfinder would be used in clinical practice.  For
example, Pathfinder is expected to be used by community hospital pathologists to
assist them in diagnosing challenging lymph-node cases.
Currently, when a community hospital pathologist gets a difficult case, he refers the
case to an expert, such as the expert on the Pathfinder project.  Therefore, the
cases selected for this experiment were chosen from a large library of cases referred
to our expert from community pathologists. Relatively old cases (older than four
months) were selected to decrease the chance that the memory of the expert would bias
the results.  

Twenty-six cases were selected at random from the referral library
such that no two diagnoses were the same.  Repeat diagnoses were not
allowed so that the inference methods would be evaluated over a larger
portion of the lymph node knowledge base.  To account for the fact
that some diseases are much more likely to occur than others, the
ratings derived from the metrics for each case are weighted by the
relative likelihood of occurrence of the case.  The relative
likelihoods were computed by normalizing the prior probabilities of
the true diagnosis of each case so that they summed to one.\footnote{In
pathology, several methods are used to establish a true diagnosis.
In some cases, a diagnosis is established through the use of expensive
tests.  In other cases, a diagnosis is established through observation of the
time course of a patient's illness.  In still other cases, a diagnosis
can be established only by an expert pathologist examining tissue
sections under a microscope.  In this study, all three approaches,
including combined approaches, were used.}

Although the cases were selected at random, a postexperiment analysis showed that the
cases were more challenging than a set of average cases would be.  The expert
reported that 50 percent of the cases contained many more technical imperfections
(such as tears and poor preservation) than is usual.  He also thought that 70 percent
of the cases were more difficult to diagnose than the average case.  The deviation
from normal probably occurred because the case-selection process
favored the inclusion of rare diagnoses.

A pathology resident entered the observations for each case into a computer database
after examining lymph-node biopsies through a microscope. A pathology resident was
used for two reasons.  First, our expert could not be allowed to look at the lymph
nodes slides directly, because he would observe more information than is presented to
the inference methods.  In addition, the expertise of a resident closely matches the
expertise of the users targeted for Pathfinder.

The manner in which features were selected for identification
deviated from the approach typically used in Pathfinder.
Specifically, a pathologist usually enters only a few salient
features and then receives recommendations from Pathfinder about
what additional features are most useful for narrowing the diagnostic
contenders.  The pathologist then provides values for one or more
of these recommended features, and the process cycles.  To avoid
confounding the performance of the inference methods with that of the feature
recommendation strategies, the resident was asked to enter all
``salient features observed.''  At no time was the resident allowed to
see what features the system recommended to be evaluated.

Once features values had been identified for each case, they were presented to the
three inference methods, producing three belief distributions.  The expert was then
given two evaluation sheets for each case.  The first sheet included a list of the
observations identified by the resident, as well as list of all the disease
hypotheses represented in Pathfinder.  The expert was asked to assign a probability
distribution to the diseases based on the observations given.  The descriptive gold
standard was derived from this distribution.  The second sheet was identical to the
first, except that it included the distributions produced by the three inference
methods.  The distributions were displayed in columns in a random order for each
case.  The expert was asked to rate each belief distribution using the 0 to 10 scale
described earlier, and again to assign a probability distribution to the diseases. 
He was allowed to refer to his first probability distribution during the second
assignment.  The informed gold standard was derived from this second distribution.

In two of the twenty-six cases, the expert found the lists of observations
confusing.  Also, in these same two cases, the simple Bayes and odds--likelihood
inference methods produced inconsistent distributions in which all hypotheses were
assigned a belief of zero.  Consequently, these two cases were removed from the study.

\section{Results}

Decision-theoretic ratings for five different procedures for determining a diagnosis
are shown in Table \ref{results1}.  ``Informed gold standard'' refers to the
procedure of prescribing the disease that maximizes utility under the distribution
used to derive the informed gold standard (Equation \ref{eval1}).   ``Simple
Bayes-MEU'' refers to the procedure of prescribing the disease that maximizes utility
under the simple Bayes distribution.   ``Simple Bayes,'' ``Odds--likelihood,'' and
``Dempster--Shafer'' refer to the procedures of prescribing the most likely diseases
under the simple Bayes, odds--likelihood, and Dempster-Shafer distributions,
respectively.

The values in the first column represent the absolute decrease in
utility of a patient when faced with the result of a lymph-node biopsy and
diagnosed using a particular approach.  The values represent an
average over the 26 cases examined, weighted by the
likelihood of occurrence of each case.  Notice that most of the
decrease in each case is attributed to the fact that the patient is
sick.  Errors in diagnosis account for little of the decrease in
utility.  In particular, the rating associated with the informed gold
standard represents the decrease in utility associated with the best
possible diagnosis under the conditions of the experiment and
therefore reflects solely the decrease in utility of the patient due
to illness.  This rating shows that a patient with a lymph-node
biopsy faces a decrease in utility of 205,804 micromorts, on average.
That is, the patient is as bad off as he would be facing a 0.2 chance of
immediate, painless death.  This quantity dominates the decreases in
utility due to diagnostic error.

To highlight the effects of diagnostic error, differences between the
informed gold standard rating and the rating for each diagnostic
approach are shown in column 2 of Table \ref{results1}.  The
standard deviation of these differences is given in column 3 of
the table.  Note that the standard deviations are quite large
relative to the mean differences.  The reason for such large
variances is easily appreciated.  For each diagnostic approach, the
diagnosis prescribed by the approach is identical to the diagnosis
prescribed by the gold standard in many of the 24 cases.  In
particular, the simple Bayes-MEU, simple Bayes, and odds--likelihood
approaches prescribe the gold-standard diagnosis in 17 of 24
cases.  The naive Dempster--Shafer approach prescribes the gold-standard diagnosis in
12 of 24 cases.  In these cases, the ratings for the gold standard and diagnostic
approaches are equal. In the remaining cases, the approaches prescribe a diagnosis
that differs from the gold standard prescription.  These nonoptimal
diagnoses are often associated with utilities that are significantly
lower than is the utility associated with the gold-standard diagnosis.
Thus, differences in utility fluctuate from zero in many cases to
large values in others, resulting in large standard deviations.

Although the standard deviations are high, a Monte Carlo permutation
test indicates that the performance of the naive Dempster--Shafer
approach is significantly inferior to that of the other methods (achieved
significance level = 0.004).  No other significant difference exists
among the other methods.

\begin{table*}
\begin{center}
\begin{tabular}{rcccc}
 & \multicolumn{4}{c}{Decision-theoretic ratings (micromorts)} \\
                            &               &       & \multicolumn{2}{c}{Differences} \\
                            & Absolute mean &       & mean   & sd \\ \cline{2-5}
    Informed gold standard  &  205,804      &       &  --    & -- \\
          Simple Bayes-MEU  &  206,615      &       & 811    & 4079 \\
              Simple Bayes  &  206,635      &       & 831    & 4078 \\
           Odds--likelihood &  206,635      &       & 831    & 4078 \\
     Naive Dempster--Shafer &  216,371      &       & 10,587 & 19,101
\end{tabular}
\end{center}
\caption{Decision-theoretic ratings of the inference methods.}
\label{results1}
\end{table*}

The expert ratings for each inference method are shown in Table
\ref{results2}.  As in the decision-theoretic approach, the mean and
standard deviation are weighted by the relative prior probability of
the true diagnosis.  Because the ratings apply directly to the belief
distributions derived by each method, there is no distinction between
the simple Bayes-MEU and simple Bayes procedures.

Using the expert-rating metric, another significant difference is
detected.  In particular, a Wilcoxon two-sample rank test shows that
the simple Bayes inference procedure performs significantly better than
does the odds--likelihood approach (achieved significance level = 0.07).

\begin{table*}
\begin{center}
\begin{tabular}{rcc}
                             & \multicolumn{2}{c}{Expert ratings} \\
                             & mean   &   sd     \\  \cline{2-3}
               Simple Bayes  & 8.52   &  1.17    \\
            Odds--likelihood & 7.33   &  1.95    \\
      Naive Dempster--Shafer & 0.03   &  0.17    
\end{tabular}
\end{center}
\caption{Expert ratings of the inference methods.}
\label{results2}
\end{table*}

Table \ref{results3} shows a comparison of the informed
and descriptive gold standards.  The differences between the two
standards are not significant.  Thus, seeing the belief distributions
generated by the inference methods did not persuade the expert to
change his opinion about the cases to any significant degree.  Of
course, this finding should not be generalized to other experts or
to other domains.

\begin{table*}
\begin{center}
\begin{tabular}{rcccc}
 & \multicolumn{4}{c}{Decision-theoretic ratings (micromorts)} \\
                            &               &       & \multicolumn{2}{c}{Differences} \\
                            & Absolute mean &       & mean   & sd \\ \cline{2-5}
    Informed Gold Standard  &  205,804      &       &  --    & -- \\
 Descriptive Gold Standard  &  205,888      &       & 84     & 1273 \\
\end{tabular}
\end{center}
\caption{Decision-theoretic ratings of expert distributions.}
\label{results3}
\end{table*}

\section{Discussion}

Before examining the results in detail, it is useful to make some general comments
about the two evaluation metrics.  An obvious advantage of the decision-theoretic
approach over the expert-rating approach is that its results are much more
meaningful.  For example, the difference between the simple Bayes and naive
Dempster--Shafer ratings using the expert-rating metric is 8.5 on a scale from 0 to
10 and is deemed to be ``significant'' by a standard statistical test.  The
difference of approximately 10,000 micromorts between the two approaches as
determined by the decision-theoretic metric, however, carries much more force; it
implies that using the naive Dempster--Shafer approach instead of the simple Bayes
approach is equivalent to assuming an additional one in 100 risk of death!  

A disadvantage of the decision-theoretic with respect to the expert-rating approach
is that its results have limited scope. Specifically, the differences among inference
methods may be highly dependent on the assessments of diagnostic utility made by our
expert.  Furthermore, decision-theoretic comparisons of inference methods are likely
to vary from one domain to another because there is room for wide variation in utility
assessments between domains.  The results of the experimental comparison must be
considered in this light.

An advantage of the expert-rating metric over the decision-theoretic
metric, as demonstrated in this experiment, is that the former can be
much more sensitive to differences.  For example, the
decision-theoretic ratings of the simple Bayes and of the
odds--likelihood methods are identical.  In contrast, the expert-rating
metric shows the two inference methods to be significantly different.
High sensitivity is likely to be a property of the expert-rating
approach across many domains.  In a typical consulting session,
an expert is hypersensitive to errors in diagnosis, whether such errors matter
to a decision maker or not, because the integrity of the expert is on the line.
It is likely that this hypersensitivity will carry over into
expert-rating ratings of diagnostic performance.  This advantage of
using an expert-rating metric is not absolute.  Considerations of
integrity or liability, for example, can always be incorporated into
the diagnostic utilities.  Indeed, the fact that components of
preference can be made explicit and are under the direct control of
the expert is one advantage of the decision-theoretic approach.

Another advantage of the expert-rating metric is that it is less
time-consuming to implement.  It took the expert approximately 20
hours, working with two people trained in decision analytic
techniques, to develop the utility model used in this evaluation.  It
took the expert less than 1 minute per case to rate the 
distributions produced by the three inference methods.

Overall, the two approaches are complementary.  The expert-rating approach is useful
for identifying differences in performance that may be important in {\em some}
domain.  The decision-theoretic metric reveals the degree of importance of such
differences for a particular domain of interest.  
It should be mentioned that information-theoretic metrics exist for measuring
differences between probability distributions, such as relative entropy and the Brier
score \cite{Ben-Bassat78} \cite{Spiegelhalter86}. The advantages and disadvantages of
the  information-theoretic and expert-rating methods are similar with respect to the
decision-theoretic approach, except that the information-theoretic methods require
probabilistic interpretations for the distributions to be compared.

Given these considerations about the evaluation metrics, differences in performance
among the inference methods can now be discussed.  In this experimental comparison,
the method for selecting an optimal diagnosis with the highest decision-theoretic
rank is simple Bayes-MEU.  The difference between the rank of this method and the
gold standard is 811 micromorts.  With the caveats described previously, this value
can be seen to represent the maximum room for improvement in the knowledge base.  Such
improvements may include more careful assessments of probabilities in the knowledge
base, and the representation of dependencies among features.

The difference in ratings between simple Bayes-MEU and simple Bayes
is only 20 micromorts and is not significant.  This result suggests that,
in the lymph-node domain, little is gained by using the more
sophisticated decision rule.  Three factors of this domain appear to
be responsible for this observation.  First, the
resident pathologist recorded all salient features observed under the
microscope for each case.  Second, the
lymph-node domain appears to be structured such that, when all
salient features are entered, most of the probability mass will fall
on one disease hypothesis or a set of disease hypotheses within the
same utility equivalence class.  In 20 of the 24 cases,
95 percent of the probability mass fall on a set of diseases within
the same equivalence class. 
Third, the structure of diagnostic utilities in the domain is such that a
disease with small probability will rarely be chosen as the optimal
diagnosis using the principle of maximum expected utility.  In light
of these factors, the relative value of decision rules should not be
extrapolated to other domains without explicit justification.

Several interesting observations can be made about the relative performances of the
simple Bayes and odds--likelihood inference methods. First, the expert-rating metric
shows a significant difference between these methods, whereas the decision-theoretic
metric shows no difference between them. This result is a clear example of the
decreased sensitivity of the decision-theoretic approach to evaluation.

Second, the theoretical difference between the simple Bayes and
odds--likelihood inference methods is that the former assumes evidence
to be conditionally independent on the hypotheses, as shown in
Equation \ref{ciH}, whereas the latter assumes evidence to be
conditionally independent on both the hypotheses and on the negation of
the hypotheses, as reflected in Equations \ref{ciH} and \ref{cinotH}.
Thus, the decision-theoretic and expert-rating results show that,
although the additional assumption of conditional independence on the
negation of hypotheses is inconsequential in the lymph-node domain,
it may lead to significant degradation in performance in other
domains.

Third, there is a regularity in the differences between the
distributions produced by the two methods.  Specifically, the
simple Bayes distributions produced in this study are, with only one
exception, more peaked.  That is, the variance of these distributions
are smaller than are those produced using the odds--likelihood approach.  This
difference can be traced to the additional assumption of conditional
independence on the negation of hypotheses, Equation \ref{cinotH}.  To
see this connection, consider a hypothetical example in which there are three
mutually exclusive and exhaustive hypotheses---$H_{1}$, $H_{2}$,
and $H_{3}$---that have equal prior probabilities.  Suppose there are
many pieces of evidence relevant to these hypotheses such that each
piece of evidence $E_{j}$ has the same probability of occurrence for
a given hypothesis.  That is, \mbox{$p(E_{j}|H_{i}) = p(E|H_{i})$}
for all $E_{j}$, and $i = 1, 2, 3$.  Also suppose that the
likelihoods have values such that:
\begin{displaymath}
p(E|H_{1})  >  p(E|H_{2})  >  p(E|H_{3})
\end{displaymath}
and
\begin{displaymath}
\begin{array}{ccccc}
\frac{p(E|H_{1})}{p(E|\overline{H_{1}})} & = & \frac{2 p(E|H_{1})}{p(E|H_{2}) + p(E|H_{
3})} & > & 1 \\
& & & & \\
\frac{p(E|H_{2})}{p(E|\overline{H_{2}})} & = & \frac{2 p(E|H_{2})}{p(E|H_{1}) + p(E|H_{
3})} & > & 1 \\
& & & & \\
\frac{p(E|H_{3})}{p(E|\overline{H_{3}})} & = & \frac{2 p(E|H_{3})}{p(E|H_{1}) + p(E|H_{
2})} & < & 1
\end{array}
\end{displaymath} 
These constraints are satisfied easily (for example,
$p(E|H_{1}) = 0.8$, $p(E|H_{2}) = 0.6$, and $p(E|H_{3}) = 0.2$).  Under these
conditions, evidence $E$ is confirmatory for $H_{1}$, confirmatory to a lesser
degree for $H_{2}$, and disconfirmatory for $H_{3}$.  Using the simple Bayes
inference procedure (Equation \ref{Bayes}) it can be shown that, as the number of
pieces of evidence grows, the posterior probability of $H_{1}$ tends to one
whereas the posterior probability of both $H_{2}$ and $H_{3}$
tends to zero.  However, using the odds--likelihood approach (Equation
\ref{BU1}) where evidence is conditionally independent given the
negation of hypotheses, a different result is obtained.  In
particular, as the number of pieces of evidence grows, it can be shown
that the posterior probabilities of both $H_{1}$ and $H_{2}$ tend
to one, whereas the posterior probability of $H_{3}$ tends to zero.
In the odds--likelihood approach, these probabilities are renormalized,
so the probabilities of $H_{1}$ and $H_{2}$ each approach
one-half.  Thus, in this example, the odds--likelihood distribution is
less peaked than is the simple Bayes distribution.  In general,
simple Bayes distributions will be more peaked, because this method
tends to amplify differences in likelihoods, whereas the odds--likelihood
method tends to washout differences.

Unlike previous observations, this one does not appear to be tied to
the lymph-node domain.  Provided a large body of evidence is reported
such that the simple Bayes approach produces a sharp distribution, the
odds--likelihood inference method should, in general, produce
distributions that are less peaked.  An important consequence of this
phenomenon is that degradation in performance due to the incorrect
assumption of conditional independence on the negation of hypotheses
is likely to occur in other domains.

A final observation about the simple Bayes and odds--likelihood inference methods is
that there is a regularity among the exceptional cases (5 of 24) in which
distributions produced by odds--likelihood were preferred to the those produced by
simple Bayes. Although obvious dependencies among features were captured by a
clustering technique, subtle ones remained unrepresented in the lymph-node knowledge
base.  It seems that the failure to represent the more subtle dependencies led to
decreased performance of the simple Bayes method relative to the odds--likelihood
method.  In particular, the incorrect assumption of conditional independence in the
simple Bayes approach led to overcounting of evidential support.  This overcounting,
in turn, produced distributions that were overly peaked.  In the odds--likelihood
approach, the impact of evidence was also overcounted. However, it appears that such
overcounting was partially compensated by the washout effect described.

The performances of the odds--likelihood and naive Dempster--Shafer
approaches are also interesting to compare.  Both evaluation metrics
revealed a significant difference between the two methods.  There are
two major theoretical differences between the inference procedures, one or
both of which may be responsible for the differences in performance.
First, from Equation \ref{ES4}, it is clear that each mass assignment
in the inference method contains a component proportional to the prior
probability of diseases.  Thus, when the masses for many different
observations are combined, the prior probability components will be
overcounted.  Priors are not overcounted in the odds--likelihood
approach.  Second, due to the way mass is assigned in the naive
Dempster--Shafer approach, disconfirmatory observations for disease
hypotheses are not recognized.  For example, if some observation
completely rules out a disease hypothesis in the odds--likelihood
method, the Dempster--Shafer mass for the disease--observation pair is
zero.  In the naive Dempster--Shafer inference method, a zero mass leaves
the score of a hypothesis unchanged.  Therefore, a hypothesis
ruledout by an observation in the odds--likelihood approach is left
with its degree of belief unchanged in the naive Dempster--Shafer approach.  It
is suspected that this difference is more significant than is the
overcounting of priors.

\section{Future Work}

The combination of the decision-theoretic
and expert-rating approaches to performance evaluation provides
useful insights about various components of the inference process
within the lymph node domain and about the inference process in
general.  This same approach to evaluation can be used to probe many
different aspects of the construction of an expert system.
For example, the Pathfinder research team has developed a set of
procedures that recommends additional features for observation to the
pathologist-user.  The methods discussed in this paper should prove
useful in evaluating the merits of these procedures.  In addition,
the Pathfinder group is currently exploring different techniques for
constructing consensus knowledge bases that combine the beliefs of
two or more experts.  Again, the evaluation methods can be used to
quantify the value of each approach.  In yet another study,
sensitivity to assessment errors in the knowledge base could be
examined.

It is hoped that the presentation of these evaluation methods will
encourage other researchers to evaluate a wide variety of issues surrounding the
building of real-world expert systems.

\section{Acknowledgments}

I thank Eric Horvitz for his assistance with the assessment of the diagnostic
utilities; Bharat Nathwani, the expert on the Pathfinder project, for his
contributions to the construction and evaluation of the system; and Doyen Nguyen for
her painstaking work to identify features under the microscope.  Eric Horvitz and
Gregory Cooper critiqued a draft of this manuscript.  Support for this work was
provided by the National Library of Medicine under grant R01-LM04529, by the National
Science Foundation under grant IRI-8703710, and by NASA-Ames.  A portion of the
computing resources was provided by the SUMEX-AIM resource under NIH grant RR-00785.

\bibliographystyle{plain}
\bibliography{compare}

\end{document}